\pdfoutput=1

\documentclass[11pt]{article}

\usepackage[preprint]{acl}

\usepackage{times}
\usepackage{latexsym}

\usepackage[T1]{fontenc}

\usepackage[utf8]{inputenc}

\usepackage{microtype}

\usepackage{inconsolata}

\usepackage{graphicx}

\usepackage{amsmath}
\usepackage{booktabs}
\usepackage{amsfonts}
\usepackage{enumerate}
\usepackage{lipsum}
\usepackage{enumitem}
\usepackage{multirow}
\usepackage{pifont}
\usepackage{hyperref}

\newcommand{\swt}[1]{{\color{green} #1}} 

\title{Sliding Window Attention Training for Efficient Large Language Models}

\author{
  \textbf{Zichuan Fu\textsuperscript{1,\thanks{Work was conducted during the internship of Zichuan Fu at Tencent YouTu Lab.}}},
  \textbf{Wentao Song\textsuperscript{2}},
  \textbf{Yejing Wang\textsuperscript{1}},
  \textbf{Xian Wu\textsuperscript{3}},
  \textbf{Yefeng Zheng\textsuperscript{3,4}},\\
  \textbf{Yingying Zhang\textsuperscript{3}},
  \textbf{Derong Xu\textsuperscript{1,5}},
  \textbf{Xuetao Wei\textsuperscript{6}},
  \textbf{Tong Xu\textsuperscript{5}},
  \textbf{Xiangyu Zhao\textsuperscript{1,\thanks{Corresponding author.}}},
\\
\\
  \textsuperscript{1} City University of Hong Kong
  \textsuperscript{2} Xi'an Jiaotong University \\
  \textsuperscript{3} Jarvis Research Center, Tencent YouTu Lab 
  \textsuperscript{4} Westlake University \\
  \textsuperscript{5} University of Science and Technology of China \\
  \textsuperscript{6} Southern University of Science and Technology
\\
  \small{
   \href{mailto:zc.fu@my.cityu.edu.hk}{zc.fu@my.cityu.edu.hk},
   \href{mailto:xy.zhao@cityu.edu.hk}{xy.zhao@cityu.edu.hk}
  }
}

\begin{document}

\maketitle
\begin{abstract}
Recent advances in transformer-based Large Language Models (LLMs) have demonstrated remarkable capabilities across various tasks. However, their quadratic computational complexity concerning sequence length remains a significant bottleneck for processing long documents. As a result, many efforts like sparse attention and state space models have been proposed to improve the efficiency of LLMs over long sequences. 
While these approaches achieve efficiency, they often require complex architectures and parallel training techniques.
This calls for a simple yet efficient model that preserves the fundamental Transformer architecture. 
To this end, we introduce \textbf{SWAT}, which enables efficient long-context handling via \textbf{S}liding \textbf{W}indow \textbf{A}ttention \textbf{T}raining. 
Specifically, SWAT replaces softmax with the sigmoid function for efficient information compression and retention. Then it utilizes balanced ALiBi and Rotary Position Embedding to stabilize training process. 
During inference, SWAT maintains linear computational complexity through sliding window attention while preserving model performance, achieving state-of-the-art (SOTA) results on eight commonsense reasoning benchmarks compared to mainstream linear recurrent architectures.
Code is available at \href{https://github.com/Fzkuji/swat-attention}{this link}.

   
    

\end{abstract}

\section{Introduction}
\label{introduction}

Large Language Models (LLMs) have demonstrated remarkable capabilities across various tasks, from text generation to complex reasoning~\cite{intro}. 
Unlike humans, who can efficiently process long contexts with memory, LLMs struggle to handle them due to quadratic complexity~\cite{Longformer}.
Despite their impressive performance on standard NLP tasks, this quadratic complexity poses a fundamental challenge for practical applications. The increasing need for efficient long-context processing, coupled with the computational constraints of current architectures, creates a pressing need for more scalable solutions.

Several approaches have been proposed to handle long sequences efficiently. These methods can be broadly categorized into two types: (1) sparse attention mechanisms~\cite{Longformer}, which reduce computation by selectively calculating the attention score, and (2) sequence models with recurrent architectures, such as linear attention variants~\cite{lineartransformer} and state space models~\cite{mamba}, which aim to process sequences efficiently through recursive hidden states.
However, these solutions face a fundamental dilemma---they either compromise model performance to achieve efficiency or propose new complex architectures that cannot fully exploit existing techniques for convenient implementation and deployment.
However, existing LLM solutions for handling long sequences often require complex architectures and parallel training techniques, making implementation and deployment more challenging, which calls for an efficient approach based on the existing Transformer architecture.

\begin{figure*}[ht]
    \hfill
    \includegraphics[width=0.95\linewidth]{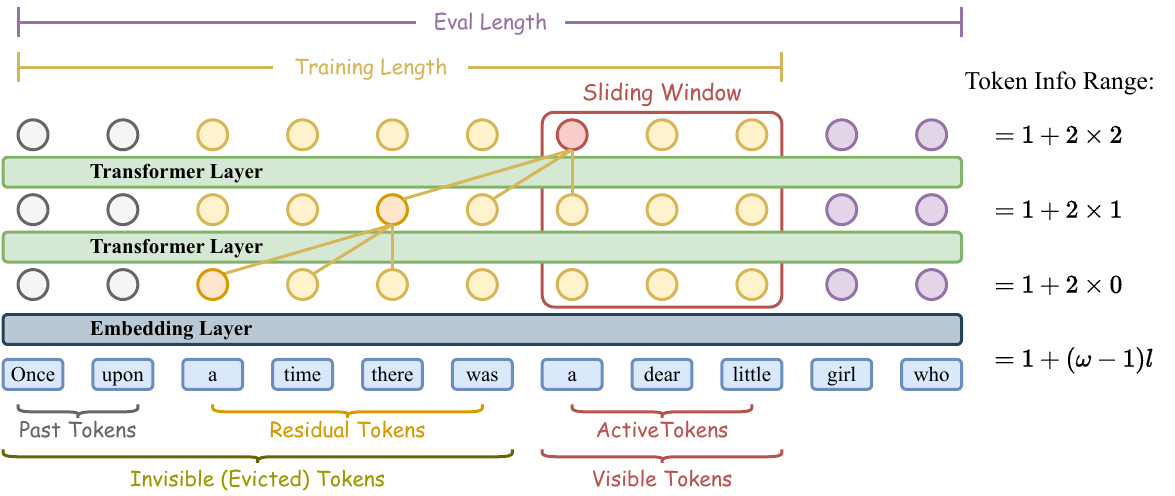}
    \caption{The demonstration of the SWA mechanism in Transformers.}
    \label{fig:swa}
\end{figure*}


Sliding Window Attention (SWA), a typical sparse attention approach~\cite{sparsetransformer}, is the most intuitive solution, as it avoids adding additional model components and compresses the inference computational complexity to linear. 
However, this approach still faces the following challenges\footnote{More details are in Section~\ref{ssec:why}}: 
(1) Current researches on SWA predominantly focus on solving the attention sink problem within the inference phase, where models allocate excessive attention to initial tokens, causing an uneven distribution of attention weights across the sequence~\cite{streamingllm}. However, they leave the training process unchanged, thereby creating a gap between inference and training.
(2) Tokens outside the attention window coverage are ignored for prediction, leading to information loss in long-context modeling~\cite{lm-infinite,sigmoidatt}.
Hence, it is crucial to investigate SWA training methods to bridge the training-inference gap and enable the model to learn long-context dependencies.

This paper introduces the SWAT framework to achieve effective SWA training and solve the aforementioned problems. Specifically, SWAT replaces the softmax operation with the sigmoid function, which not only prevents the attention sink problem but also maintains dense attention weights for higher information capacity per token.
To compensate for the lack of sparsity in sigmoid-based attention, SWAT incorporates balanced ALiBi~\cite{alibi} to introduce position-dependent differentiation, preventing information overloaded in dense representations. It also enables the model to preserve both recent and historical information effectively.
Furthermore, we enhance the framework with Rotary Position Embedding (RoPE)~\cite{rope} to explicitly encode positional information in hidden states, ensuring training stability.
SWAT trained with SWA from scratch is ultimately capable of compressing arbitrarily long texts into a fixed-length hidden state of tokens while maintaining effective information processing.
Our contributions can be summarized as follows:
\begin{itemize}[leftmargin=*, itemsep=0pt]
    \item We empirically analyze the poor performance of the SWA inference and attribute this to the attention sink problem caused by the high variance of softmax operation.
    \item We introduce SWAT, which combines sigmoid activation with balanced position embeddings, enabling effective information preservation and achieving SWA training.
    \item Extensive experiments confirm that SWAT surpasses vanilla Transformer and other recurrent models, achieving strong performance across tasks with linear computational complexity.
\end{itemize}

\section{Understanding Transformer's Attention}

\begin{figure*}[ht]
    \hfill
    \includegraphics[width=\linewidth]{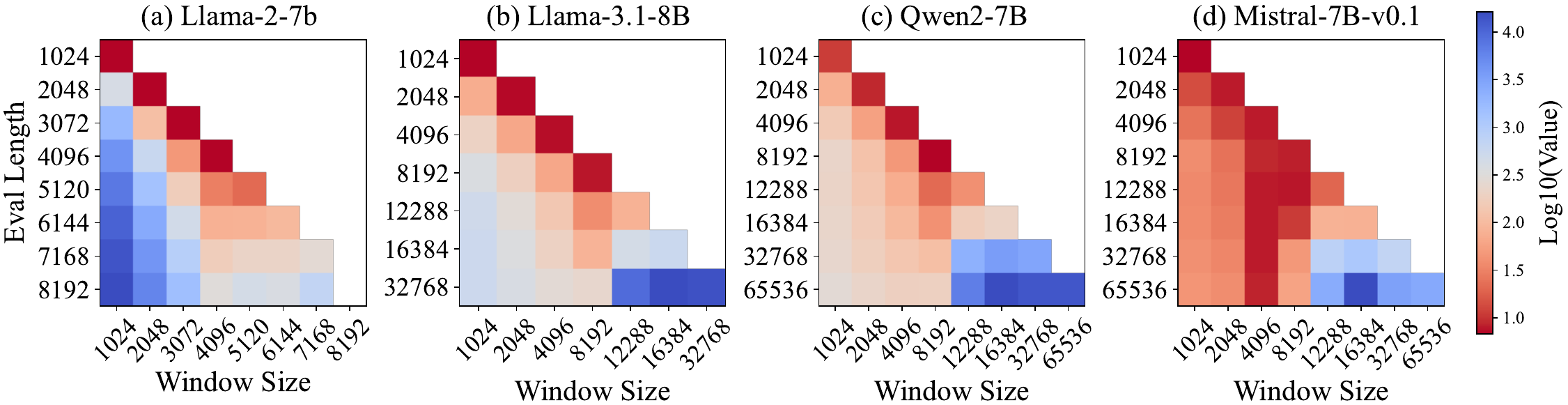}
    \caption{The $\log_{10}$ perplexity of four LLMs (Llama-2-7b, Llama-3.1-8B, Qwen2-7B and Mistral-7B-v0.1) on the third book of PG-19 test set using SWA inference. The window sizes are set not to exceed their respective training sequence lengths. The x-axis represents the sliding window size, and the y-axis represents the evaluation sequence length. For a fixed window size, perplexity increases (color shifts to blue) as the evaluation length grows.}
    \label{fig:open-llms}
\end{figure*}

\begin{figure*}[ht]
    \centering
    \includegraphics[width=\linewidth]{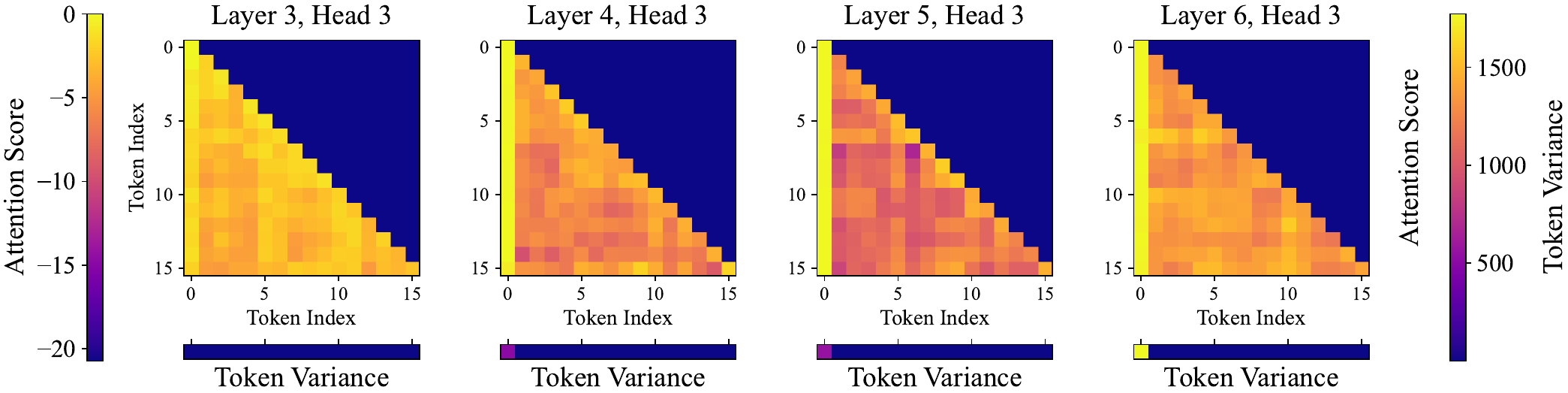}
    \caption{Heatmaps of attention scores (top four squares) and token embedding variance (bottom four lines) across different layers of Qwen2-7B. Higher token variance corresponds to stronger attention, highlighting their correlation. The two color bars indicate respective scales.}
    \label{fig:variance}
\end{figure*}

This section introduces concepts of the SWA mechanism and its potential capability in handling long sequences. We then analyze why current LLMs with SWA inference fail to achieve the expected theoretical advantages. 

\subsection{Sliding Window Attention}

The self-attention layer in Transformers typically has $O(N^2)$ computational complexity, where $N$ is the input sequence length. 
To reduce this complexity while preserving the sequential information, sliding window attention (SWA) is introduced in Longformer~\cite{Longformer}. 
SWA restricts each token to only attend the attention calculation of its neighboring tokens within a fixed-size window.
With a window size of $\omega \ll N$, the computation cost per token is reduced to $O(\omega)$, leading to an overall linear complexity $O(N \cdot \omega)$, which is more efficient than vanilla attention.

We visualize the SWA mechanism in Figure~\ref{fig:swa}, where the window size is three ($\omega=3$) and the depth is two ($L=2$).
We define the tokens that are visible to the current window as active tokens (the red block in the figure, corresponding active tokens are ``a dear little'').
For invisible tokens, also referred to as evicted tokens, we further categorize them as residual and past tokens. 
Residual tokens are not visible to the sliding window at the embedding layer. However, their information will passed to the neighboring $\omega -1$ tokens with a transformer layer (this information transition is represented as yellow lines in the figure), thus partially preserved for the prediction. For example, the information of the token `a' (the orange ball at the embedding layer) can be retained in the other token `a' (the red ball at the second transformer layer) in our visualization. Theoretically, the information range of a single token at the $l^{th}$ transformer layer is $1+(\omega-1) \cdot l$ and the maximum range is $1+(\omega-1) \cdot L$, i.e., $1+2\cdot2=5$ in the figure.



\subsection{LLMs with SWA Inference}
\label{ssec:why}

Although current open-source LLMs are structurally capable of conducting SWA inference, they fail to achieve stable improved results. As shown in Figure~\ref{fig:open-llms}, we analyzed the perplexity (PPL) of four open-source LLMs~\cite{llama2,llama3,mistral-7b,qwen2} using different sliding window sizes on the PG-19~\cite{pg19} test set. The experimental results reveal that these LLMs achieve optimal performance only when operating within their training sequence length. For instance, for Llama-2-7b model in Figure~\ref{fig:open-llms}(a), when the window size is fixed at 1,024, the perplexity gradually increases as the evaluation length grows, as indicated by the color transition from blue to red in the heatmap.
This suggests that Transformers inherently learn contextual patterns specific to their training length and fail to extend to variable-length texts during inference.

We suggest that this failure can be attributed to two major issues:
(1) the attention sink phenomenon, where models become overly dependent on initial tokens, 
and (2) information loss that past tokens are discarded.

The attention sink phenomenon~\cite{streamingllm}, where LLMs allocate excessive attention to initial tokens in sequences, has emerged as a significant challenge for SWA inference in Transformer architectures. Previous work has made two key observations regarding this phenomenon. First, the causal attention mechanism in Transformers is inherently non-permutation invariant, with positional information emerging implicitly through token embedding variance after softmax normalization~\cite{variance}. Second, studies have demonstrated that removing normalization from the attention mechanism can effectively eliminate the attention sink effect~\cite{whensinkemerge}.



Based on these insights, we analyze the attention patterns and hidden state statistics of Qwen2-7B, as shown in Figure~\ref{fig:open-llms}. Our results reveal a strong correlation between token variance and attention sink magnitude---the variance of hidden states for the first token is significantly higher than for subsequent tokens. \textit{This finding provides strong evidence that attention sink manifests through variance propagation via normalization.} Notably, even though models like Qwen2 incorporate explicit relative position embeddings (e.g., RoPE), they still learn and rely on this implicit absolute positional information through the normalization mechanism.

Beyond the attention sink problem, softmax also leads to significant information loss during sliding window inference. Consider the following example of how softmax transforms attention scores:
\begin{equation}
\begin{bmatrix}
1.5 \\
5.0 \\
2.4 \\
0.5 \\
1.3
\end{bmatrix}
\to \text{Softmax}(x_i) = \frac{e^{x_i}}{\sum_{j} e^{x_j}} \to
\begin{bmatrix}
0.03 \\
0.88 \\
0.07 \\
0.01 \\
0.02
\end{bmatrix}
\end{equation}
As shown above, the exponential nature of softmax dramatically amplifies differences between logits, causing most of the probability mass to concentrate on the highest-scoring token (0.88 in this case) while severely suppressing other tokens (all below 0.07). A detailed mathematical proof of this sparsification property is provided in Appendix~\ref{app:sparsity}.

In summary, while softmax's sparsification is beneficial for full-context Transformers, it becomes limiting in SWA scenario where the aggressive filtering impedes the model's ability to retain historical information within the sliding window.


\section{Sliding Window Attention Training}

In this section, we explore the advantages of SWA training over traditional Transformer training with a new paradigm for processing long sequences. Additionally, we provide a detailed explanation of our proposed SWAT attention layer. This simple yet effective attention layer combines Sigmoid~\cite{sigmoid}, ALiBi, and RoPE to address the information retention challenges of SWA.

\subsection{Information Transmission}

\begin{figure}[t]
    \centering
    \includegraphics[width=\linewidth]{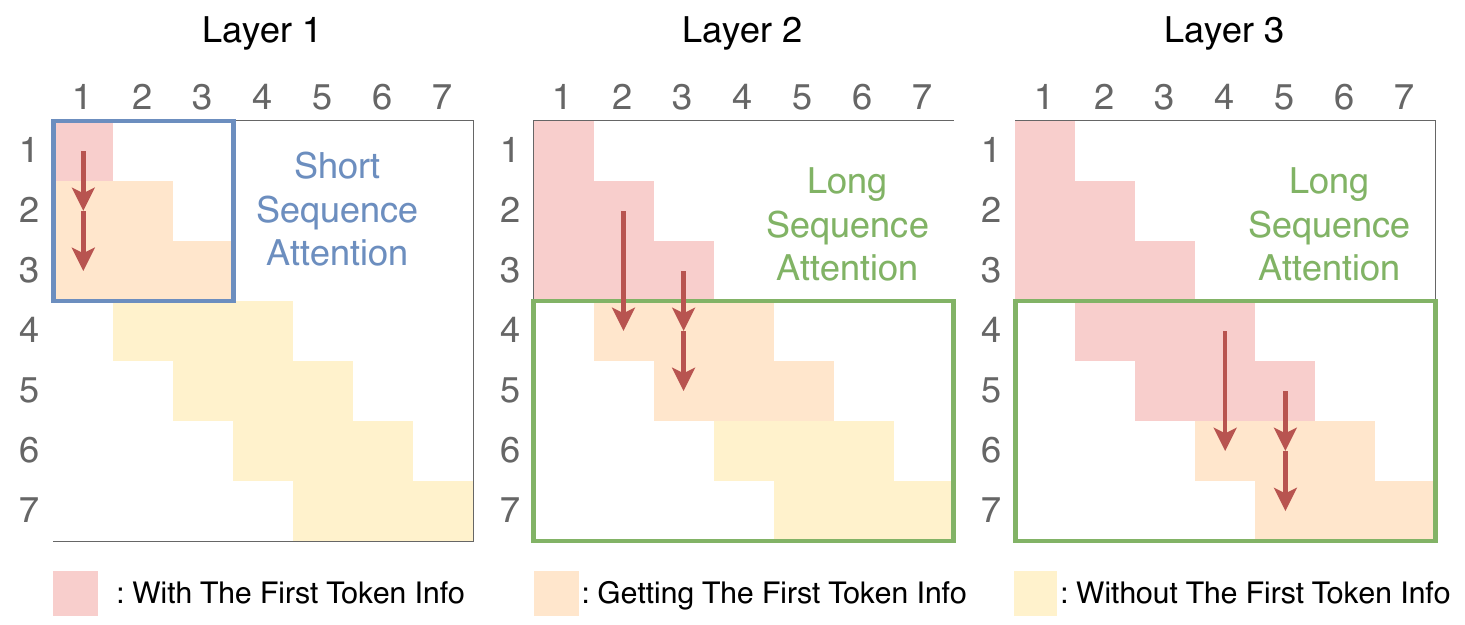}
    \caption{The demonstration of the SWA mechanism in Transformers, where the model's information coverage includes residual and active tokens, depending on the model depth and window size.}
    \label{fig:att}
\end{figure}

Traditional Transformer training involves processing entire sequences of tokens, allowing the model to capture long-range dependencies through global attention mechanisms. In contrast, SWA operates within a limited context, necessitating new approaches to preserve information continuously. As shown in Figure~\ref{fig:att}, SWA training enables two distinct learning paradigms for LLMs, short and long sequence attentions.

In conventional Transformer training, the sequence length is smaller than the window size. New tokens can acquire and integrate information from all tokens, even the very first tokens in the text. Therefore, the model keeps essential information in each token embedding and enhances the ability to extract information, which is also strengthened by the softmax function.

SWA training introduces a new training paradigm, where each window shift requires careful historical context management. In particular, the old token embedding is discarded after sliding. However, in the upper layers of the Transformer, the new token's embedding still retains the old token's embedding with a certain weight. Hence, the model tends to retain all past embeddings in the upper-level model to prevent information loss caused by sliding windows, strengthening the model's ability to compress information. The experimental results demonstrating how SWA training enhances the model's capabilities are presented in Sections~\ref{ssec:swat} and \ref{ssec:ablation}.

\subsection{Attention Computation}
\label{ssec:Attention-Computation}

In this subsection, we propose SWAT, a modified attention mechanism that combines sigmoid activation with integrated position embeddings. The input consists of queries, keys, and values with dimension of $d$. Instead of using softmax normalization, we apply sigmoid activation to the scaled dot products to obtain attention weights, preventing mutual suppression between tokens:
\begin{equation}
\text{Attention}(\boldsymbol{Q}, \boldsymbol{K}, \boldsymbol{V}) = \sigma (\frac{\boldsymbol{Q}\boldsymbol{K}^T}{\sqrt{d}})\boldsymbol{V}
\end{equation}
where $\boldsymbol{Q} \in \mathbb{R}^{N \times d}$, $\boldsymbol{K} \in \mathbb{R}^{N \times d}$, and $\boldsymbol{V} \in \mathbb{R}^{N \times d}$ are packed matrices of queries, keys, and values, respectively; $\sigma ( \cdot )$ is the sigmoid function. More detailed analysis can be found in Appendix~\ref{app:density}.

To introduce discriminative bias in the dense attention patterns of sigmoid activation and better differentiate token representations within sliding windows, we propose balanced ALiBi, a bidirectional extension of the original ALiBi mechanism. For an input subsequence within a window, we add position-dependent biases to the attention scores:
\begin{equation}
\text{Attention}(\boldsymbol{Q}, \boldsymbol{K}, \boldsymbol{V}) = \sigma (\frac{\boldsymbol{Q}\boldsymbol{K}^T}{\sqrt{d}} + s \cdot (m-n))\boldsymbol{V}
\end{equation}
where $m$ and $n$ ($m >le n$) denote the index of tokens in the sequence and $s$ denotes the slope.
Unlike the original ALiBi, which uses only negative slopes to enforce a directional inductive bias, we use both positive and negative slopes across different attention heads. For a model with $h$ heads, we assign positive slopes to $h/2$ heads and negative slopes to the remaining heads. The magnitude of slopes follows a geometric sequence similar to ALiBi, but in both directions:
\begin{equation}
s_k = \begin{cases}
-2^{-k} & \text{for forward-looking heads} \\
2^{-k} & \text{for backward-looking heads}
\end{cases}
\label{eq:-+}
\end{equation}
where $k$ ranges from 1 to $h/2$ for each direction. This bidirectional slope design allows attention heads to specialize in different temporal directions, with forward-looking heads focusing on recent context and backward-looking heads preserving historical information.

After replacing softmax with sigmoid, the implicit position information through normalization is lost, leading to training instability. Furthermore, while balanced ALiBi provides positional variance through attention weights, its positional signals remain weak. To address this issue, we further incorporate RoPE to enhance explicit positional information. Finally, SWAT attention calculates the attention output as follows:
\begin{equation}
    \begin{aligned}
        & \text{Attention}(\boldsymbol{Q}, \boldsymbol{K}, \boldsymbol{V})_m = {\textstyle \sum_{n=m-\omega+1}^{m}}  \\
        & \sigma \Bigg( 
        \frac{(\boldsymbol{R}_{\Theta, m}^d \boldsymbol{q}_m)^T (\boldsymbol{R}_{\Theta, n}^d \boldsymbol{k}_n)}
        {\sqrt{d_k}}  \quad + s \cdot (m-n) \Bigg) \boldsymbol{v}_n
    \end{aligned}
\end{equation}
where $\boldsymbol{R}_{\Theta, m}^d$ and $\boldsymbol{R}_{\Theta, n}^d$ are the same rotation matrices as Equation 15 in \cite{rope}. To ensure SWA training, note that $m-n < \omega$.

This combination of sigmoid activation, balanced ALiBi, and RoPE makes up for the sparsity of the vanilla Transformer. It ensures the stability of training and strengthens the information contained in a single token embedding.

\subsection{Network Efficiency}

Since SWAT's architecture is nearly identical to a standard attention layer, the per-token computation cost remains almost the same under an equivalent attention length—apart from the additional overhead of computing the ALiBi. However, the overall computation becomes linear due to the use of a sliding window. Thus, the inference computational complexity can be expressed as:
\begin{equation}
\mathrm{Cost} =N  \omega \times ( 1+\delta_{\text{ALiBi}}), 0 < \delta_{\text{ALiBi}} \ll 1
\end{equation}
where $\delta_{\text{ALiBi}}$ represents the extra cost of ALiBi.

\section{Experiments}
\label{experiments}

\begin{table*}[t]
\tiny
\caption{Overall comparison of SWAT and other models on eight common-sense reasoning tasks. Bold values represent optimal performance, while second-best values are underlined. ``\textbf{{ *}}'' indicates the statistically significant improvements (i.e., two-sided t-test with $p<0.05$) over the best baseline. $\uparrow$: higher is better. $\downarrow$: lower is better.}
\label{tab:overall} 
\resizebox{\textwidth}{!}{
\begin{tabular}{@{}lccccccccccc@{}}
\toprule
\multicolumn{1}{l|}{Model} & \begin{tabular}[c]{@{}c@{}}Wiki.\\ ppl $\downarrow$\end{tabular} & \multicolumn{1}{c|}{\begin{tabular}[c]{@{}c@{}}LMB. \\ ppl $\downarrow$\end{tabular}} & \begin{tabular}[c]{@{}c@{}}LMB. \\ acc $\uparrow$\end{tabular} & \begin{tabular}[c]{@{}c@{}}PIQA\\ acc $\uparrow$\end{tabular} & \begin{tabular}[c]{@{}c@{}}Hella. \\ acc\_n $\uparrow$\end{tabular} & \begin{tabular}[c]{@{}c@{}}Wino. \\ acc $\uparrow$\end{tabular} & \begin{tabular}[c]{@{}c@{}}ARC-e\\ acc $\uparrow$\end{tabular} & \begin{tabular}[c]{@{}c@{}}ARC-c\\ acc\_n $\uparrow$\end{tabular} & \begin{tabular}[c]{@{}c@{}}SIQA\\ acc $\uparrow$\end{tabular} & \begin{tabular}[c]{@{}c@{}}BoolQ\\ acc $\uparrow$\end{tabular} & \begin{tabular}[c]{@{}c@{}}Avg.\\ $\uparrow$\end{tabular} \\ \midrule \midrule
\multicolumn{12}{c}{340M params / 15B tokens} \\ \midrule
\multicolumn{1}{l|}{Transformer++} & 31.52 & \multicolumn{1}{c|}{41.08} & 30.76 & 62.98 & 34.76 & 50.53 & 45.21 & 24.05 & 36.81 & 58.24 & 42.92 \\
\multicolumn{1}{l|}{RetNet} & 32.50 & \multicolumn{1}{c|}{49.73} & 28.24 & 62.61 & 34.15 & 50.91 & 44.27 & 23.62 & 36.79 & 59.72 & 42.54 \\
\multicolumn{1}{l|}{GLA} & 28.51 & \multicolumn{1}{c|}{43.02} & 28.73 & 64.05 & 35.96 & 50.00 & 54.19 & 24.29 & 37.13 & 58.39 & 44.09 \\
\multicolumn{1}{l|}{Mamba} & 30.83 & \multicolumn{1}{c|}{40.21} & 29.94 & 63.79 & 35.88 & 49.82 & 49.24 & 24.56 & 35.41 & 60.07 & 43.59 \\
\multicolumn{1}{l|}{DeltaNet} & 28.65 & \multicolumn{1}{c|}{47.30} & 28.43 & 63.52 & 35.95 & 49.63 & 52.68 & 25.37 & \underline{37.96} & 58.79 & 44.04 \\
\multicolumn{1}{l|}{TTT} & 27.44 & \multicolumn{1}{c|}{34.19} & 30.06 & 63.97 & 35.71 & 50.08 & 53.01 & 26.11 & 37.32 & 59.83 & 44.51 \\
\multicolumn{1}{l|}{Gated DeltaNet} & \underline{27.01} & \multicolumn{1}{c|}{\underline{30.94}} & \underline{34.11} & 63.08 & 38.12 & \underline{51.60} & 55.28 & 26.77 & 34.89 & 59.54 & 45.42 \\
\multicolumn{1}{l|}{Titans} & \textbf{26.18} & \multicolumn{1}{c|}{\textbf{29.97}} & \textbf{34.98} & 64.73 & \textbf{39.61} & \textbf{51.85} & 55.60 & \underline{28.14} & 34.52 & 59.99 & \underline{46.17} \\
\multicolumn{1}{l|}{SWAT (-)} & 33.32 & \multicolumn{1}{c|}{36.75} & 32.80 & \textbf{ 65.94*} & \underline{38.99} & 50.12 & \textbf{ 59.68*} & \textbf{ 28.24*} & \textbf{ 38.69*} & \underline{60.55} & \textbf{ 46.88*} \\
\multicolumn{1}{l|}{SWAT (+)} & 37.47 & \multicolumn{1}{c|}{49.15} & 29.59 & 65.40 & 36.92 & 50.43 & 54.55 & 26.88 & 37.67 & 58.93 & 45.05 \\
\multicolumn{1}{l|}{SWAT (-+)} & 35.53 & \multicolumn{1}{c|}{45.06} & 29.96 & \underline{65.67} & 37.39 & 50.91 & \underline{56.99} & 27.05 & 36.75 & \textbf{ 62.11*} & 45.85 \\ \midrule
\multicolumn{12}{c}{760M params / 30B tokens} \\ \midrule
\multicolumn{1}{l|}{Transformer++} & 25.21 & \multicolumn{1}{c|}{27.64} & 35.78 & 66.92 & 42.19 & 51.95 & 60.38 & 32.46 & 39.51 & 60.37 & 48.69 \\
\multicolumn{1}{l|}{RetNet} & 26.08 & \multicolumn{1}{c|}{24.45} & 34.51 & 67.19 & 41.63 & 52.09 & 63.17 & 32.78 & 38.36 & 57.92 & 48.46 \\
\multicolumn{1}{l|}{Mamba} & 28.12 & \multicolumn{1}{c|}{23.96} & 32.80 & 66.04 & 39.15 & 52.38 & 61.49 & 30.34 & 37.96 & 57.62 & 47.22 \\
\multicolumn{1}{l|}{Mamba2} & 22.94 & \multicolumn{1}{c|}{28.37} & 33.54 & 67.90 & 42.71 & 49.77 & 63.48 & 31.09 & 40.06 & 58.15 & 48.34 \\
\multicolumn{1}{l|}{DeltaNet} & 24.37 & \multicolumn{1}{c|}{24.60} & 37.06 & 66.93 & 41.98 & 50.65 & 64.87 & 31.39 & 39.88 & 59.02 & 48.97 \\
\multicolumn{1}{l|}{TTT} & 24.17 & \multicolumn{1}{c|}{23.51} & 34.74 & 67.25 & 43.92 & 50.99 & 64.53 & \underline{33.81} & \textbf{40.16} & 59.58 & 47.32 \\
\multicolumn{1}{l|}{Gated DeltaNet} & \underline{21.18} & \multicolumn{1}{c|}{22.09} & 35.54 & 68.01 & 44.95 & 50.73 & \textbf{66.87} & 33.09 & 39.21 & 59.14 & 49.69 \\
\multicolumn{1}{l|}{Titans} & \textbf{20.04} & \multicolumn{1}{c|}{21.96} & 37.40 & 69.28 & \underline{48.46} & 52.27 & \underline{66.31} & \textbf{35.84} & \underline{40.13} & \textbf{62.76} & \underline{51.56} \\
\multicolumn{1}{l|}{SWAT (-)} & 23.41 & \multicolumn{1}{c|}{\underline{21.05}} & \textbf{ 40.81*} & \textbf{ 69.80*} & \textbf{ 48.65*} & 51.69 & 65.15 & 33.53 & 39.95 & 61.07 & \textbf{ 51.85*} \\
\multicolumn{1}{l|}{SWAT (+)} & 23.91 & \multicolumn{1}{c|}{\textbf{21.05}} & 39.01 & 69.59 & 47.64 & \underline{53.43} & 64.73 & 32.34 & 39.15 & 57.95 & 50.48 \\ 
\multicolumn{1}{l|}{SWAT (-+)} & 23.34 & \multicolumn{1}{c|}{21.36} & \underline{39.08} & \underline{69.70} & 48.16 & \textbf{53.91*} & 65.15 & 31.06 & 39.41 & \underline{61.62} & 51.01 \\
\bottomrule
\end{tabular}
}
\end{table*}

\subsection{Experiment Settings}

\paragraph{Datasets.}

For the overall comparison, models are trained on the 100BT subset of FineWeb-Edu~\cite{fineweb-edu}, which is a high-quality educational dataset designed for LLM pre-training.



\paragraph{Baselines.}

Our baselines include state-of-the-art models including both vanilla Transformer and recurrent models. Specifically, we compare our approach against Transformer++~\cite{llama2}, RetNet~\cite{retnet}, Gated Linear Attention (GLA)~\cite{gla}, Mamba~\cite{mamba}, DeltaNet~\cite{deltanet}, TTT~\cite{ttt}, Gated DeltaNet~\cite{gateddeltanet}, and Titans~\cite{titans}.

\paragraph{Implementation Details.}

We pre-train SWAT with model sizes of 340M and 760M parameters on 15B and 30B tokens, respectively. The training uses the same vocabulary as Llama 2~\cite{llama2}, with a sequence length of 4096 tokens and a batch size of 0.5M tokens.


\paragraph{Evaluation Metrics.}

We evaluate model performance using perplexity (ppl), accuracy (acc), and normalized accuracy (acc\_n). Perplexity measures language modeling ability, where lower values indicate better predictions. Accuracy assesses classification performance by calculating the proportion of correct predictions. Normalized accuracy is adjusts for dataset difficulty variations, ensuring fair comparisons across different evaluation settings.

\begin{table*}[t]
\caption{Performance comparison of language models pretrained with and without sliding windows.}
\label{tab:performance_comparison}  
\resizebox{\textwidth}{!}{
\begin{tabular}{@{}l|ccc|cccc|cccc|c@{}}
\toprule
\multirow{2}{*}{\textbf{Models}} & \multirow{2}{*}{\textbf{\begin{tabular}[c]{@{}c@{}}Training\\ Window\end{tabular}}} & \multirow{2}{*}{\textbf{\begin{tabular}[c]{@{}c@{}}Training \\  Length\end{tabular}}} & \multirow{2}{*}{\textbf{\begin{tabular}[c]{@{}c@{}}Eval\\ Window\end{tabular}}} & \multicolumn{4}{c|}{\textbf{OpenWebText (Eval Length=)}} & \multicolumn{4}{c|}{\textbf{PG-19 (Eval Length=)}} & \textbf{OpenOrca} \\ \cmidrule(l){5-13} 
 &  &  &  & 128 & 1,024 & 4,096 & 16,384 & 128 & 1,024 & 4,096 & 16,384 & - \\ \midrule
Vanilla A & 128 & 128 & 128 & \textbf{3.2490} & 3.6536 & 3.6761 & 4.8414 & 4.9682 & 5.2139 & 5.1529 & 5.6949 & 6.0084 \\
Sliding Window A & 128 & 1,024 & 128 & 3.3619 & 3.1286 & 3.0766 & 3.0051 & 5.1785 & 4.8164 & 4.7510 & 4.7663 & 7.7471 \\
Vanilla B & 1,024 & 1,024 & 128 & 3.3395 & 3.3042 & 3.2856 & 3.2379 & 5.6052 & 5.0742 & 5.0797 & 5.1336 & 7.9706 \\
Vanilla B & 1,024 & 1,024 & 1,024 & 3.3395 & \textbf{2.9716} & \textbf{2.9541} & 2.9636 & 5.6052 & 5.3429 & 5.1517 & 5.0274 & 7.9706 \\
Vanilla B & 1,024 & 1,024 & 16,384 & 3.3395 & \textbf{2.9716} & 3.5534 & 3.0786 & \textbf{3.3395} & \textbf{2.9716} & 5.4912 & 5.2372 & 7.9706 \\
Sliding Window B & 1,024 & 4,096 & 1,024 & 3.4380 & 3.0197 & 2.9638 & \textbf{2.9128} & 5.0880 & 4.6587 & 4.5107 & \textbf{4.4383} & \textbf{5.8802} \\
Vanilla C & 4,096 & 4,096 & 4,096 & 3.3788 & 2.9784 & 2.9705 & 2.9518 & 5.1519 & 4.5444 & \textbf{4.4366} & 4.4938 & 5.9315 \\
Vanilla D (Upper Bond) & 16,384 & 16,384 & 16,384 & \multicolumn{4}{c|}{OOM} & \multicolumn{4}{c|}{OOM} & OOM \\ \bottomrule
\end{tabular}
}
\end{table*}

\begin{table*}[t]
\caption{Performance comparison of language models with different activation functions and position embeddings.}
\label{tab:table3}  
\resizebox{\textwidth}{!}{
\begin{tabular}{@{}l|c|ccccc|cccc@{}}
\toprule
\textbf{No.} &
  \textbf{\begin{tabular}[c]{@{}c@{}}Model \\ Type\end{tabular}} &
  \textbf{\begin{tabular}[c]{@{}c@{}}Activation\\ Function\end{tabular}} &
  \textbf{\begin{tabular}[c]{@{}c@{}}Position\\ Embedding\end{tabular}} &
  \textbf{\begin{tabular}[c]{@{}c@{}}Training\\ Window\end{tabular}} &
  \textbf{\begin{tabular}[c]{@{}c@{}}Training \\ Length\end{tabular}} &
  \textbf{\begin{tabular}[c]{@{}c@{}}Eval\\ Window\end{tabular}} &
  \textbf{OpenWebText} &
  \textbf{PG-19} &
  \textbf{OpenOrca} &
  \textbf{Avg.} \\ \midrule
1  & Vanilla & Softmax & RoPE        & 128  & 128  & 128  & 4.8414          & 5.6949          & 6.0085          & 5.5149          \\
2  & Vanilla & Sigmoid & RoPE        & 128  & 128  & 128  & 14.2562         & 15.4765         & 1.9906          & 10.5744         \\
3  & Sliding & Softmax & RoPE        & 128  & 1,024 & 128  & 3.0140          & 4.7839          & 6.9671          & 4.9217          \\
4  & Sliding & Sigmoid & ALiBi-12:0  & 128  & 1,024 & 128  & 3.0073          & 4.6895          & 0.1631          & 2.6200          \\
5  & Sliding & Sigmoid & ALiBi-8:4   & 128  & 1,024 & 128  & 3.0391          & 4.6435          & 0.2650          & 2.6492          \\
6  & Sliding & Sigmoid & ALiBi-6:6   & 128  & 1,024 & 128  & 3.0484          & 4.9920          & \textbf{0.1420} & 2.7275          \\
7  & Sliding & Sigmoid & ALiBi-6:6   & 128  & 2,048 & 128  & 3.0634          & 5.0384          & 0.1712          & 2.7577          \\
8  & Sliding & Sigmoid & AliRope-6:6 & 128  & 1,024 & 128  & 3.0486          & \textbf{4.3103} & 0.1709          & \textbf{2.5099} \\
9  & Sliding & Sigmoid & AliRope-6:6 & 1,024 & 1,024 & 1,024 & 2.9716          & 4.3915          & 0.5304          & 2.6312          \\
10 & Vanilla & Softmax & RoPE        & 1,024 & 1,024 & 1,024 & \textbf{2.9631} & 4.5447          & 5.4702          & 4.3260          \\
11 & Vanilla & Sigmoid & ALiBi       & 1,024 & 1,024 & 1,024 & 2.9659          & 5.0681          & 0.1717          & 2.7352          \\ \bottomrule
\end{tabular}
}
\end{table*}

\subsection{Overall Performance}


In this section, we evaluate the performance of SWAT on eight commonsense reasoning benchmarks, as detailed in Appendix~\ref{app:benchmarks}. The comparison is conducted on 340M and 760M parameter models. 
For our SWAT, (-) denotes negative slopes (i.e., the negative ALiBi slope to look forward in Equation~\ref{eq:-+}); (+) denotes positive slopes, which use the opposite slope of ALiBi (i.e., the positive slope in Equation~\ref{eq:-+} looking backward); and (-+) indicates that half of the attention heads have negative slopes and half have positive slopes. 

As shown in Table~\ref{tab:overall}, SWAT (-) achieves state-of-the-art  (SOTA) performance on average (46.88\%) across eight common sense reasoning tasks, surpassing all other baselines. This is mainly attributed to the short-text benchmarks, such as PIQA and Hellaswag, where SWAT (-) focuses more on the information from newly input tokens.
Although SWAT (-) initially shows higher perplexity than other baselines at 340M parameters, when scaled to 760M parameters, it demonstrates strong decreases in perplexity on Wiki and LMB. This suggests a performance improvement trend for larger models with the sigmoid function.
On the contrary, the purely forward-looking SWAT (+) shows weaker performance, suggesting that forward slopes work best combined with backward attention. 

The balanced configuration SWAT (-+), where attention heads are evenly split between looking forward and backward, achieves more uniform performance across different tasks by effectively processing both recent and historical information. Specifically, SWAT (-+) achieves the best performance (62.11\%) on BoolQ, a question-answering dataset where historical context is crucial for accurate predictions. This result aligns with our findings in Section~\ref{ssec:ablation}, where balanced attention heads demonstrate superior performance on both OpenOrca and PG-19 datasets, confirming the importance of balanced historical information processing for complex reasoning tasks. Meanwhile, due to the allocation of some attention heads for remembering information from older tokens, SWAT (-+) shows a slight performance compromise on shorter benchmarks. However, this issue is alleviated as the model scales from 340M to 760M.
The results remain consistent at 760M parameters, showing robustness across model sizes.

\subsection{Sliding Window Attention Training}
\label{ssec:swat}

To verify the effectiveness of SWA training, we conduct experiments comparing vanilla Transformers pre-trained with and without SWAT training across three datasets. Using Llama2-based models~\cite{llama2} pretrained on OpenWebText, we investigate the impact of varying sliding window sizes and sequence lengths, with results shown in Table~\ref{tab:performance_comparison}. In the table, vanilla Transformers are which training length are the same as their training window size, and the labels A, B, C, and D represent the model identifiers. 

When the sliding window mechanism is applied, we observe a notable improvement in performance, particularly with longer evaluation sequence lengths. For instance, in the Sliding Window A configuration, when the evaluation length is 16,384, Sliding Window A achieves a performance of 3.0051 on OpenWebText, surpassing the 4.8414 achieved by Vanilla A. Additionally, Sliding Window B achieves the best performance across all three datasets when the evaluation length is 16,384. Note that all results are from models trained for 80,000 steps. If training continues, the attention sink issue is likely to worsen, further degrading vanilla model performance.

Based on our experimental results, we draw two key conclusions: 
(1) Wtih the same model structure, SWA training significantly improves performance, especially with longer evaluation sequence lengths. This is likely because SWA training forces the model to retain memory of older information across long sequences, while vanilla models struggle with memory as they retain all historical tokens.
(2) The vanilla Transformers perform optimally only when the evaluation length matches the training length, whereas the SWA trained models maintain consistent performance across varying sequence lengths. This is likely because vanilla Transformers heavily attend to initial tokens due to attention sink, while SWA models learn to focus primarily on the current window, ensuring stable performance across different sequence lengths.

\begin{figure}[t]
    \centering
    \includegraphics[width=\linewidth]{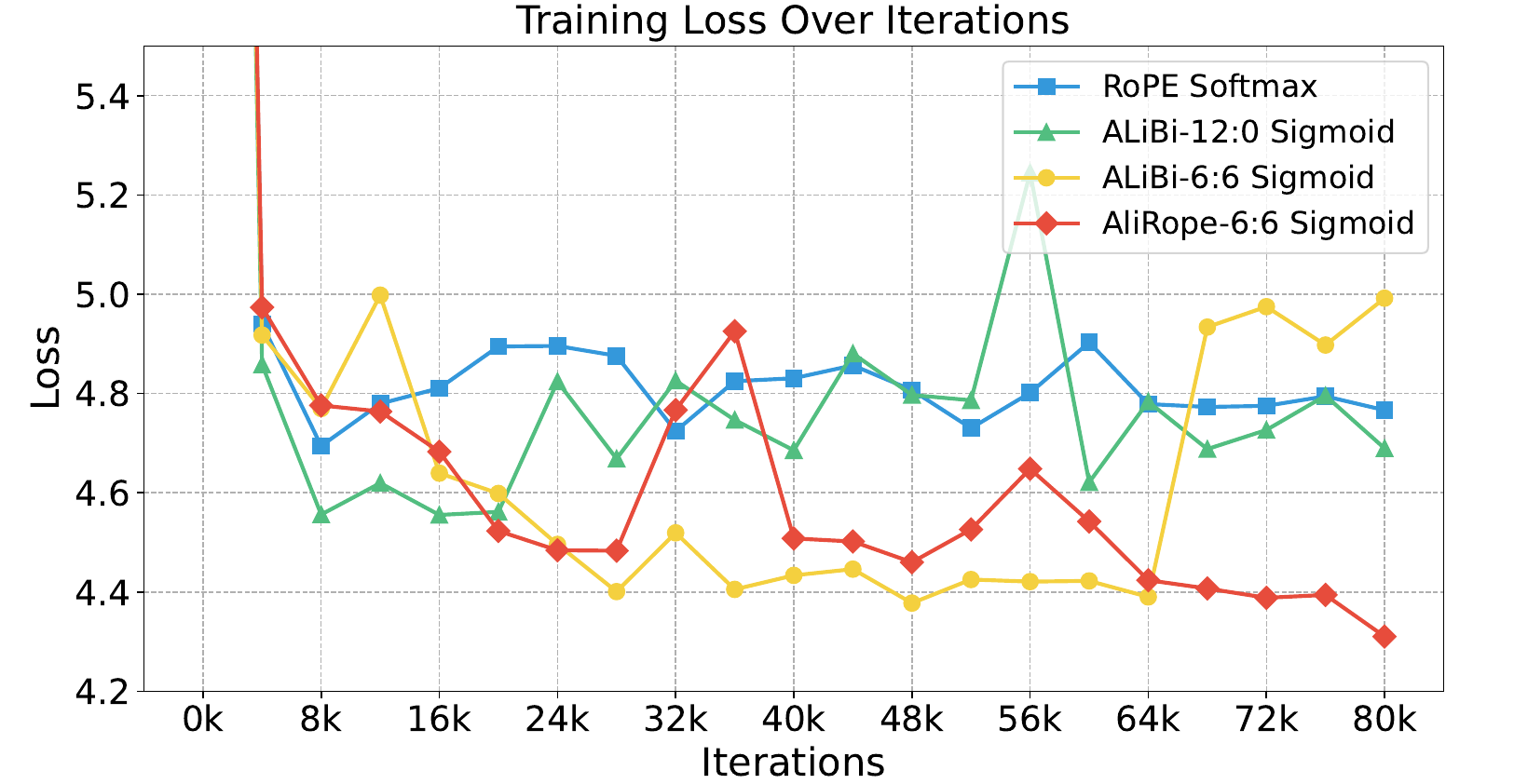}
    \caption{The training loss of models with different modules including Sigmoid, RoPE, and ALiBi, with the balanced slopes.}
    \label{fig:loss}
\end{figure}

\subsection{Ablation Study}
\label{ssec:ablation}

This section evaluates the impact of activation functions, position embeddings, and ALiBi slopes.
We systematically test 11 different configurations (No.1-11) to understand how different combinations of model components affect long-context performance, as shown in Table~\ref{tab:table3} and Figure~\ref{fig:loss}.

Comparing No.1 and No.2, directly replacing softmax with sigmoid in vanilla Transformer leads to significant performance degradation, likely due to overloaded information in token embeddings without mutual suppression. However, using ALiBi stabilizes training by distinguishing subtle differences in token embeddings based on position information (No.10 and No.11). Furthermore, the slope configuration plays a key role, with No.5 and No.6 outperforming No.4, suggesting a better balance between recent and past information. However, Figure~\ref{fig:loss} shows that training instability persists at later stages (ALiBi-6:6 Sigmoid), indicating that ALiBi alone provides weak positional information. AliRope-6:6 Sigmoid (No.8) achieves the lowest loss values among all variants, with 2.51 on average, while demonstrating more stable training pattern as shown in Figure~\ref{fig:loss}. Finally, comparing No.7 and No.6, extending the training length from 1,024 to 2,048 while keeping the number of layers and window size fixed does not help with the loss.

\section{Related Works}
\label{related-works}

\subsection{Efficient Transformers}

While architectural innovations offer one path to efficiency, research also focuses on optimizing the Transformer itself, particularly through sparse attention patterns to reduce computational cost.

Early work in this direction focused on structured sparsity patterns. Sparse Transformer~\cite{sparsetransformer} demonstrated that using fixed sparse attention patterns could maintain model performance while significantly reducing computation. This idea was further developed by Longformer~\cite{Longformer} and BigBird~\cite{bigbird}, which introduced more sophisticated attention patterns combining local windows with global tokens to capture dependencies effectively. 
These models, however, still rely on predefined attention patterns, which can limit flexibility. \swt

\subsection{Efficient LLMs}

To address the quadratic complexity of Transformers, researchers have proposed various efficient models categorized into the following categories:

\textbf{Linear Recurrent Models} achieve $O(n)$ complexity through different approximation techniques. Linear Transformer~\cite{lineartransformer} replaces softmax attention with kernel functions, while Performer~\cite{performers} employs random feature approximation. Recent works like GLA~\cite{gla} introduce forgetting mechanisms to prevent information explosion, while Gated Delta Networks~\cite{gateddeltanet} focus memory updates to enable both precise memory updates and quick resets when needed. Models like Mamba~\cite{mamba} and RWKV~\cite{rwkv} take a fundamentally different approach by utilizing state space models (SSMs) instead of attention, providing an alternative way to capture sequential patterns.

\textbf{Memory-Augmented Architectures} enhance Transformers' ability to handle long sequences by incorporating explicit memory mechanisms. For example, Transformer-XL~\cite{transformer-xl} pioneered the use of cached computations from previous segments with relative positional embeddings. More recent works like Memorizing Transformers~\cite{memorizingtransformers} and Focused Transformer~\cite{focusedtransformer} try to store and retrieve relevant historical information.

While these models achieve better efficiency, their complex architectures often lead to more challenging optimization compared to standard Transformers, which benefit from simple and well-established training procedures.



\section{Conclusion}
\label{sec:conclusion}

This paper introduces SWAT, a new architecture for efficient LLMs via sliding window attention training, which maintains the core Transformer architecture. By replacing softmax with sigmoid and combining balanced ALiBi with RoPE, SWAT addresses the attention sink issue and ensures stable training. SWAT enables effective information compression and retention across sliding windows without complex architectural changes. Experimental results show that SWAT outperforms other models across eight common-sense reasoning benchmarks, excelling in tasks that require long-range comprehension. Future work could explore adaptive window sizes for more flexible text processing.

\section{Limitations}

While our architectural design ensures relatively robust training stability, SWAT's performance exhibits significant sensitivity to hyperparameter configuration. Critical parameters including window size, model depth, and the distribution of ALiBi slopes substantially impact model efficacy. This necessitates comprehensive hyperparameter exploration to optimize the model architecture.

Additionally, as the model scales, it may encounter diminishing returns in retaining long-context information. In particular, larger models may fully memorize training data, reducing the need for information transmission, which in turn weakens the effectiveness of mechanisms designed to handle extended contexts. Future experiments will need to keep cache from previous steps during training to address this problem.

Finally, despite SWAT's strong overall performance, the model exhibits an inherent limitation in its attention mechanism. Specifically, SWAT's maximum attention distance is constrained by the product of window size and model depth. Although extending these parameters can theoretically increase the attention span, information loss remains inevitable when processing ultra-long sequences. For applications requiring complete information retention over extensive contexts, alternative approaches such as hybrid architectures or explicit memory retrieval mechanisms may be necessary to complement SWAT's capabilities.


\bibliography{custom}

\appendix

\section{Why Does the Softmax Function Lead to Sparsity?}
\label{app:sparsity}

In models such as Transformers, dot-product attention is the most widely used approach. Let a query vector $\boldsymbol{q}$ and multiple key vectors $\boldsymbol{k}_1, \boldsymbol{k}_2, \ldots, \boldsymbol{k}_L$ be given, where $\boldsymbol{q}, \boldsymbol{k}_i \in \mathbb{R}^d$. We stack the key vectors into a matrix:
\begin{equation}
\boldsymbol{K} \;=\;
\begin{bmatrix}
\boldsymbol{k}_1 \\
\boldsymbol{k}_2 \\
\vdots \\
\boldsymbol{k}_L
\end{bmatrix}.
\end{equation}
The attention distribution (i.e., the set of attention weights) $\boldsymbol{\alpha}$ is computed by:
\begin{equation}
\boldsymbol{\alpha} = \text{softmax} \left (\tfrac{\boldsymbol{q} \boldsymbol{K}^\top}{\sqrt{d}} \right ),
\end{equation}
where $\text{softmax}(z_i) = e^{z_i} / \sum_j e^{z_j}$. Let
\begin{equation}
E_i = \frac{\boldsymbol{q}\cdot \boldsymbol{k}_i}{\sqrt{d}},
\end{equation}
so the $i$-th attention weight is:
\begin{equation}
\alpha_i = \frac{\exp(E_i)}{\sum_{j=1}^n \exp(E_j)}.
\end{equation}

Sparsity arises because the exponential function greatly amplifies any $E_i$ that is larger than the rest: if $E_1$ is significantly bigger than $E_2, \dots, E_L$, then $\exp(E_1)$ will dominate the sum in the denominator, pushing $\alpha_1$ close to $1$ and making the others near $0$. Formally, define
\begin{equation}
\Delta_i = E_1 - E_i \quad \text{for } i \ge 2,
\end{equation}
so we have:
\begin{equation}
\begin{aligned}
\frac{\alpha_i}{\alpha_1} &= \frac{\exp(E_i)}{\exp(E_1)} \\
                          &= \exp (E_i - E_1 ) \\
                          &= \exp (-\Delta_i ).
\end{aligned}
\end{equation}
If $\Delta_i$ is large and positive, then $\exp(-\Delta_i)$ is very small, causing $\alpha_i$ to vanish compared to $\alpha_1$. Moreover, in high-dimensional spaces (i.e., when $d$ is large), random dot products $\boldsymbol{q} \cdot \boldsymbol{k}_i$ tend to have higher variance, making it more likely that one or a few $E_i$ values will stand out dramatically. This ``winner-takes-most'' scenario becomes amplified, thereby increasing the tendency toward sparsity within the attention distribution.

In practice, the dot-product $\boldsymbol{q} \cdot \boldsymbol{k}_i$ often yields extreme values—meaning that one or a few of the resulting energies $E_i$ are substantially larger than the others. This phenomenon causes the softmax to concentrate most of the probability mass on these extreme values. To rigorously analyze this behavior, we suppose each attention score $E_i$ is an independent and identically distributed (i.i.d.) random variable drawn from a Gaussian distribution:
\begin{equation}
E_i \sim \mathcal{N}(\mu, \sigma^2).
\end{equation}

Under this assumption, by the central limit theorem, the dot product $\boldsymbol{q}\cdot \boldsymbol{k}_i$ follows an approximately normal distribution after appropriate scaling. More importantly, extreme value theory states that the maximum value among $L$ i.i.d. Gaussian variables, denoted as $E_{(L)} = \max_{1 \le i \le L} E_i$, satisfies approximately: 
\begin{equation}
E_{(L)} \approx \mu + \sigma \sqrt{2 \ln L}.
\end{equation}
In contrast, a typical attention score is around $\mu$. Therefore, the expected gap between the maximum energy and a typical energy is on the order of: 
\begin{equation}
\Delta \approx \sigma \sqrt{2 \ln L}.
\end{equation}
Given this gap, we have:
\begin{equation}
\frac{\alpha_i}{\alpha_1} \approx \exp\Bigl(-\sigma \sqrt{2 \ln L}\Bigr). 
\end{equation}
For large $L$, this ratio becomes exponentially small.

\section{Why Does the Sigmoid Function Maintain Density?}
\label{app:density}

While the softmax function induces a probability distribution over multiple inputs, the sigmoid function operates on each input independently and does not normalize across multiple values. Concretely, the sigmoid of a scalar $z$ is defined as:
\begin{equation}
\sigma(z) \;=\; \frac{1}{1 + e^{-z}}.
\end{equation}

In contrast to softmax—which computes exponential terms for all inputs $z_1, z_2, \dots, z_L$ and divides by their sum—sigmoid only involves a single exponential term $e^{-z}$ within its own calculation. Consequently, one input's value does not directly compete with another input's value in a shared denominator. Since the final attention weight for each token is determined independently based on its relationship with the query, there is no ``winner-takes-most'' effect as seen in softmax-based attention.

Finally, in a sigmoid-based attention mechanism, the computed token embedding can retain information from all tokens within the attention window, rather than being dominated by a single token with high attention weight. To effectively preserve the diversity of token integration, it is important to ensure that the embedding dimension is sufficiently large. A higher dimensional space allows different token values to be effectively combined while maintaining meaningful distinctions between them.

\section{Detailed Experiment Settings}
\label{app:experiment-settings}

\subsection{Datasets}
While our main experiments utilize a specific high-quality educational dataset, we conducted preliminary evaluations across multiple datasets to comprehensively assess model capabilities. All datasets are split according to the ratio: train:validation:test = 8:1:1. Here we detail the characteristics and purposes of each dataset. 

Our overall experiment employs a 100 billion token subset of \textbf{FineWeb-Edu}~\cite{fineweb-edu}, which is specifically curated for language model pre-training. This dataset consists of high-quality educational content that provides well-structured training examples for developing fundamental language understanding capabilities.

\begin{table}[t]
\centering
\caption{Statistics of the datasets used in our analysis experiments. All datasets are in English and split into train, validation, and test sets with a ratio of 8:1:1. Sample sizes are reported in millions (M) or thousands (K).}
\label{tab:dataset}
\resizebox{\linewidth}{!}{
\begin{tabular}{@{}lclcccc@{}}
\toprule
\textbf{Name} & \textbf{Task} & \textbf{Usage} & \textbf{Language} & \textbf{Train} & \textbf{Validation} & \textbf{Test} \\ \midrule
OpenWebText & Language Modeling & All & English & 6.48M & 0.81M & 0.81M \\
PG-19 & Language Modeling & Test & English & 15.6M & 1.95M & 1.95M \\
OpenOrca & Question Answering & Test & English & 400K & 50K & 50K \\ \bottomrule
\end{tabular}
}
\end{table}

For our subsequent experiments, as shown in Table~\ref{tab:dataset}, we deliberately selected three complementary datasets that evaluate different aspects of model performance:

\textbf{OpenWebText}~\cite{openwebtext} comprises predominantly shorter web-based texts. It provides a foundation for assessing basic language modeling capabilities. In contrast to specialized corpora, OpenWebText's diverse content allows evaluation of general language understanding across varied domains and writing styles.

\textbf{PG-19}~\cite{pg19} is based on complete books published before 1919, presenting a distinct challenge in processing long-form literary content. The book-length texts require models to maintain coherence and compress information across extended narratives, testing their ability to capture long-range dependencies and thematic consistency. 

\textbf{OpenOrca}~\cite{OpenOrca} is a question-answering dataset that tests models' information retention capabilities. This is particularly important as the answers to questions are often embedded in earlier parts of the context, making it an effective benchmark for assessing models' ability to maintain essential information when processing long sequences.

We utilized OpenWebText for traininga and validation, while incorporating all three datasets into the test phase.
To thoroughly evaluate long-context processing capabilities, we extended the input sequence length to 16,384 tokens for both OpenWebText and PG-19. This multi-dataset evaluation framework allows us to systematically analyze model performance across different linguistic challenges and context lengths, providing a comprehensive view of their capabilities and limitations.

\subsection{Benchmarks}
\label{app:benchmarks}

For our overall experiment, we compare models on eight common-sense reasoning tasks, in Table~\ref{tab:bench}:

\textbf{Wikitext}~\cite{wikitext}: A large linguistic corpus extracted from Wikipedia articles, containing over 100 million word tokens. It tests a model's ability to predict the next word in a passage of text.

\textbf{Lambada}~\cite{lambada}: The LAmBdA dataset tests a model's capability of using broad discourse context to predict the last word of a passage extracted from books. It contains over 60,000 examples.

\textbf{PIQA}~\cite{PIQA}: The Physical Interaction: Question Answering (PIQA) dataset tests commonsense reasoning about physical interactions between two entities. It contains 16,113 multiple choice questions generated from crowd-sourcing.

\textbf{Hellaswag}~\cite{Hellaswag}: The HellaSwag dataset consists of 70,000 multiple choice questions about inferring what might happen next in a story. It requires commonsense reasoning to choose the most plausible ending.

\textbf{WinoGrande}~\cite{WinoGrande}: The WinoGrande dataset tests coreference resolution and commonsense reasoning with 44,000 examples obtained from books and websites. 

\textbf{ARC}~\cite{arc}: The AI2 Reasoning Challenge (ARC) dataset contains 7,787 genuine grade-school level, multiple-choice science questions, grouped into an Easy Set (ARC-e) and a Challenge Set (ARC-c).

\textbf{SIQA}~\cite{siqa}: The Social Interaction QA (SIQA) dataset contains 15,554 multiple choice questions that describe situations about people's social interactions. 

\textbf{BoolQ}~\cite{boolq}: The Boolean Questions (BoolQ) dataset contains 15,942 English yes/no questions sampled from Google search queries to test a model's ability to answer simple questions.

\begin{table}[t]
\centering
\caption{The statistics of the benchmarks used in the overall experiment.}
\label{tab:bench}  
\resizebox{0.8\linewidth}{!}{
\begin{tabular}{@{}l|c@{}}
\toprule
\textbf{Dataset} & \textbf{Sample Size} \\ \midrule
Wikitext & 60,634 \\
Lambada & 60,000 \\
PIQA & 16,113 \\
Hellaswag & 70,000 \\
WinoGrande & 44,000 \\
ARC & 7,787 (Easy Set + Challenge Set) \\
SIQA & 15,554 \\
BoolQ & 15,942 \\ \bottomrule
\end{tabular}
}
\end{table}

\subsection{Implementation Details.}

\paragraph{Overall Experiment}

In the overall experiment (Table~\ref{tab:overall}), SWAT means we pretrain the model with our sliding window attention training. 
We pre-train SWAT with model sizes of 340M and 760M parameters on 15B and 30B tokens, respectively. The SWAT models are compared to other language models of similar sizes.

Evaluations measure perplexity (lower is better) and accuracy (higher is better) on datasets like PIQA, WinoGrande, and BoolQ. 
For our SWAT, as defined in Equation~\eqref{eq:-+}, 
(-) denotes the configuration using only negative slopes (i.e., traditional ALiBi slopes $s_k = -2^{-k}$),
(+) denotes the configuration using only positive slopes (i.e., $s_k = 2^{-k}$),
(-+) denotes our bidirectional configuration where:
Half of the attention heads ($h/2$ heads) use negative slopes $s_k = -2^{-k}$, the other half use positive slopes $s_k = 2^{-k}$.
For both directions, $k$ ranges from 1 to $h/2$.
The experiments are based on two GitHub repositories flash-linear-attention\footnote{\href{https://github.com/Fzkuji/flash-linear-attention}{https://github.com/Fzkuji/flash-linear-attention}} and lm-evaluation-harness\footnote{\href{https://github.com/EleutherAI/lm-evaluation-harness}{https://github.com/EleutherAI/lm-evaluation-harness}}.

\paragraph{Analysis Experiments}

For analysis experiments, models are evaluated on three datasets: OpenWebText, PG-19, and OpenOrca, with the average accuracy reported. We experiment with different training window sizes, training lengths, and evaluation window sizes.  The experiments are based on two GitHub repositories nanoGPT\footnote{\href{https://github.com/karpathy/nanoGPT}{https://github.com/karpathy/nanoGPT}} and flash-linear-attention. We pre-train SWAT (248M parameters) for 80,000 steps with a batch size of 250k tokens, accumulating a total training exposure of 20B tokens, which amounts to  about 2 epochs over the pre-training corpus.

In Table \ref{tab:performance_comparison}, vanilla Transformers have a training length that matches their fixed training window size. Model A, B, C, and D are identifiers for pre-trained models with different configurations being compared. The columns in the table show different sequence length settings for each model configuration. The parameters used in the table are defined as follows::
\begin{itemize}[leftmargin=*, itemsep=0pt]
    \item Training window size means the maximum sequence length the model can process per training step. 
    \item Training length means the actual sequence length used for each training example, which may be shorter than the window size when using the vanilla Transformers. 
    \item Evaluation window means the maximum context provided to the model during evaluation to make predictions. 
    \item Evaluation length means the actual sequence length fed into the model per test example.
\end{itemize}

We compared pre-training using fixed token window sizes of 128, 1,024, and 4,096 versus using variable-length sliding windows. 
With sliding window pre-training, the model is exposed to longer token sequences during training, which helps improve evaluation perplexity. 
Using sliding windows allows longer sequences during training compared to fixed windows. This table shows that the best performance was achieved when the training sequence length is four times the training window size. Different evaluation window sizes are also tested to compare model performance given varying amounts of context.

In Table \ref{tab:table3}, we compared the performance of language models with different activation functions and position embeddings. Specifically, we study the model accuracy when using softmax and sigmoid as the activation functions. We also introduce RoPE, ALiBi, and AliRope as different position embedding methods. Note that ALiBi-12:0 represents the origin ALiBi model, which uses only negative slopes, while ALiBi-6:6 represents model uses half positive and half negative slopes across different attention heads.



\end{document}